\documentclass[table, 10pt]{article} 
\usepackage[table,xcdraw]{xcolor}
\usepackage[preprint]{rlc}

\usepackage{wrapfig}
\usepackage{amssymb}            
\usepackage{mathtools}          
\usepackage{mathrsfs}           
\mathtoolsset{showonlyrefs}     
\usepackage{graphicx}           
\usepackage{subcaption}         
\usepackage[space]{grffile}     
\usepackage{url}                

\usepackage{booktabs}
\usepackage{tabularx, makecell}
\usepackage[utf8]{inputenc} 
\usepackage[T1]{fontenc}    
\usepackage{hyperref}       
\usepackage{url}            
\usepackage{booktabs}       
\usepackage{amsfonts}       
\usepackage{nicefrac}       
\usepackage{microtype}      
\usepackage{placeins}
\usepackage{enumitem}
\usepackage{comment}
\usepackage{blindtext}
\usepackage{caption}
\usepackage{natbib}
\setcitestyle{numbers,open={[},close={]}} 
\usepackage{amssymb}
\usepackage{amsbsy}
\usepackage{upgreek}
\usepackage{multirow}
\usepackage{wrapfig}

\usepackage{siunitx}

\usepackage{soul}

\usepackage{ifthen}

\usepackage{acro} 
\DeclareAcronym{NN}{
    short = {NN},
    long  = {neural network},
    tag = {abbrev}
}

\DeclareAcronym{SCoBots}{
    short = {SCoBots},
    long  = {Successive Concept Bottlenecks Agents},
    tag = {abbrev}
}

\DeclareAcronym{RL}{
    short = {RL},
    long  = {reinforcement learning},
    tag = {abbrev}
}

\DeclareAcronym{MOC}{
    short = {MOC},
    long  = {Motion and Object Continuity},
    tag = {abbrev}
}

\DeclareAcronym{SPACE}{
    short = {SPACE},
    long  = {Spatially Parallel Attention and Component Extraction},
    tag = {abbrev}
}

\DeclareAcronym{PPO}{
    short = {PPO},
    long  = {Proximal Policy Optimization},
    tag = {abbrev}
}

\DeclareAcronym{VAE}{
    short = {VAE},
    long  = {Variational Autoencoder},
    tag = {abbrev}
}

\DeclareAcronym{XAI}{
    short = {XAI},
    long  = {eXplainable AI},
    tag = {abbrev}
}

\DeclareAcronym{ICB}{
    short = {ICB},
    long  = {interpretable concept bottleneck},
    tag = {abbrev}
}

\DeclareAcronym{OC}{
    short = {OC},
    long  = {object-centric},
    tag = {abbrev}
}
\usepackage{svg}

\newboolean{anonymized} 
\setboolean{anonymized}{true} 

\definecolor{blue}{RGB}{0,0,255}

\newcommand{\eg}{\emph{e.g.,}~} 

\newcommand{\ie}{\emph{i.e.,}~} 

\newcommand{\cf}{\emph{cf.}~}

\title{Interpretable End-to-End Neurosymbolic Reinforcement Learning Agents}

\author{\quad \ Nils Grandien$^{1}$ \\
\And \qquad Quentin Delfosse$^{1, 2}$ \\
\And \quad Kristian Kersting$^{1,3,4,5}$ \\
\And \\
\qquad \qquad \quad $^1$Computer Science Department, TU Darmstadt, Germany \\
\quad \ $^2$National Research Center for Applied Cybersecurity Darmstadt, Germany\\
\quad $^3$Hessian Center for Artificial Intelligence (hessian.AI), Darmstadt, Germany \\
\qquad \qquad \quad \ \  $^4$Centre for Cognitive Science, TU Darmstadt, Germany \\
$^5$German Research Center for Artificial Intelligence (DFKI), Darmstadt, Germany \\
\qquad \qquad correspondence to \texttt{quentin.delfosse@cs.tu-darmstadt.de}
}

\begin{document}
\begin{center}
\maketitle
\end{center}


\begin{abstract}
Deep \ac{RL} agents rely on shortcut learning, preventing them from generalizing to slightly different environments~\cite{delfosse2024hackatariatarilearningenvironments}. To address this problem, symbolic methods based on object-centric states have been developed.
In this work, we instantiate the symbolic \ac{SCoBots} framework~\citep{Delfosse2024InterpretableCB}. \ac{SCoBots} decompose \ac{RL} tasks into intermediate, interpretable representations, culminating in action decisions based on a comprehensible set of object-centric relational concepts. This architecture aids in demystifying agent decisions.
By explicitly learning to extract object-centric representations from raw states, employing object-centric \ac{RL}, and using policy distillation via rule extraction, this work places itself within the neurosymbolic AI paradigm, blending the strengths of neural networks with symbolic AI. We present the first implementation of an end-to-end trained SCoBot, which we apply to different Atari games, and separately evaluate its components. The results demonstrate the framework's potential to create interpretable and performant \ac{RL} systems, and pave the way for future research directions in obtaining end-to-end interpretable \ac{RL} agents.
\end{abstract}


\section{Introduction}
Despite ongoing advancements in the field, \acf{RL} continues to face numerous challenges. One such challenge is the sparsity of rewards~\citep{Andrychowicz2017HindsightER}, where the environment only rarely provides reward signals for the agent to learn from. A related issue is credit assignment~\citep{Raposo2021SyntheticRF, 10.5555/3666122.3666170}, which refers to the challenge of identifying the specific previous actions responsible for distant future rewards. Additionally, \ac{RL} agents are susceptible to learn misaligned goals~\citep{Langosco2021GoalMI, delfosse2024hackatariatarilearningenvironments}, which occurs when the objectives optimized by the \ac{RL} algorithm diverge from the intended goals of the system's designers.
The black box nature of current deep \ac{RL} approaches impedes the ability to address these challenges. Even though there have been attempts of shedding light into the black box via approaches from the field of \ac{XAI}~\citep{Guidotti2018ASO, 10.1613/jair.1.13200}, there is still room for improvement. The majority of the developed approaches rely on post-hoc explanations, which frequently result in a lack of faithfulness of the explanations~\citep{Hooker2018ABF, Chan2022ACS}. This makes it challenging to analyze an agent's policy.

To address the lack of interpretability, we instantiate the recently proposed \ac{SCoBots} framework~\citep{Delfosse2024InterpretableCB}. This approach uses an architecture that achieves interpretability by design. \ac{SCoBots} decompose the \ac{RL} problem via concept bottleneck models~\citep{pmlr-v119-koh20a} with intermediate interpretable representations. The final action selection operates on a set of interpretable relational concepts and uses an inherently interpretable model, in our implementation a rule set policy. \ac{SCoBots} facilitate human understanding of the agent and can, thereby, aid in the development and training of performing \ac{RL} agents.
As a side benefit, the interpretability of the model can improve trust into the \ac{RL} agent, which can be crucial for deployment in the real world. Additionally, by training the \ac{RL} algorithm on a set of relational concepts instead of a sequence of raw input images the complexity of the problem is being reduced, thereby improving sample efficiency~\citep{Yoon2023AnII}.

By instantiating the \ac{SCoBots} framework, we cover the fields of object representation learning, object-centric \ac{RL} and policy distillation.
Overall, this leads to a neurosymbolic AI system that combines the strengths of neural networks with symbolic AI. Neural networks are used for object representation learning and the initial \ac{RL} algorithm. The utilization of structured object-centric intermediate representations and the final step of transforming a neural policy into a rule set policy also renders the approach symbolic.

Previous work has led to components that could be suitable for the steps in the \ac{SCoBots} framework~\citep{Lin2020SPACE:, delfosse2023moc, zarlenga2021efficient}. However, these components have not yet been combined and the \ac{SCoBots} framework has so far only been evaluated using ground truth detection of the objects.
In this work, \textbf{we introduce the first end-to-end concept bottleneck agents that employ unsupervisedly trained components}\footnote{Code at \href{hhttps://github.com/nlsgrndn/SCoBots} {https://github.com/nlsgrndn/SCoBots}, \href{https://github.com/nlsgrndn/SCoBots-RL}{https://github.com/nlsgrndn/SCoBots-RL}}. As part of this, we evaluate the individual components that were presented in previous works.
The experimental setting, in which we evaluate the \ac{SCoBots}, is Atari games.
OCAtari~\citep{Delfosse2023OCAtariOA} provides access to many such games including ground truth detection of objects.

\section{Background}

\subsection{SCoBots}

\begin{figure}[t]
    \centering
    \includegraphics[width=1.0\linewidth]{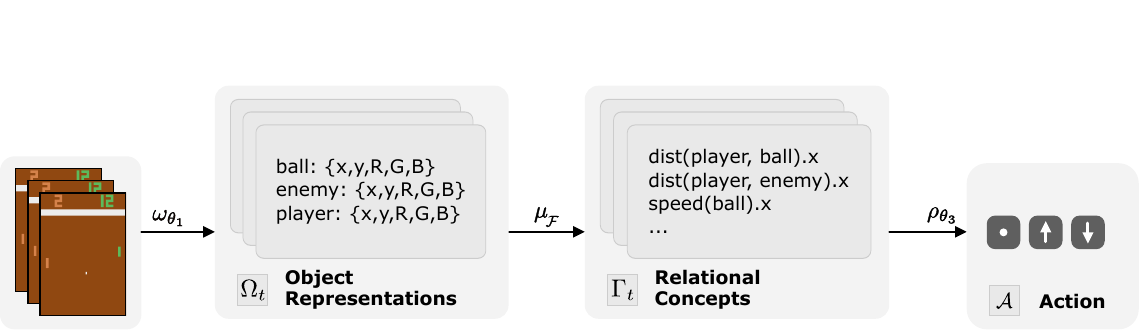}
    \caption{\textbf{Overview of the \ac{SCoBots} framework.} \ac{SCoBots} decompose the policy into three consecutive steps (\ie object extraction, relation extraction, action selection) using intermediate Interpretable Concept Bottlenecks (\acp{ICB}). This enables external users to inspect how the SCoBot agent selects its action. Figure adapted from~\citep{Delfosse2024InterpretableCB}.
    }
    \label{fig:SCoBots}
\end{figure}

In the \acf{SCoBots} framework~\citep{Delfosse2024InterpretableCB} (\cf Figure~\ref{fig:SCoBots}), the policy of an agent is decomposed into distinct steps with intermediate \acp{ICB} inspired by concept bottleneck models~\citep{pmlr-v119-koh20a}.

\begin{equation}
s_t \xrightarrow{\omega_{\theta_1}} \Omega_t \xrightarrow{\mu_{\mathcal{F}}} \Gamma_t \xrightarrow{\rho_{\theta_2}} a_t
\label{eq:SCoBots-decomposition}
\end{equation}

This differs from the standard deep \ac{RL} approach, in which the raw input is processed to directly derive the selected action without any structured intermediate steps.

\textbf{Object Extractor: $s_t \xrightarrow{\omega_{\theta_1}} \Omega_t$}
The object extractor, denoted as \(\omega_{\theta_1}(\cdot)\), extracts objects and their properties from the \(n\) most recent frames, represented by \(s_t = \{x_i\}_{i=t-(n-1)}^t\). As a result, a collection of object representations is returned: $\omega_{\theta_1}(s_t) = \Omega_t=\{o_t^j\}_{j=1}^{c_t}$. Here, \(c_i\) refers to the number of detected objects. The object representations $o^{j}_{i}$ are tensors that capture a multitude of properties of each object (\eg {\small\texttt{position}}).

\textbf{Relation Extractor: $\Omega_t \xrightarrow{\mu_{\mathcal{F}}} \Gamma_t$}
In this step, relational concepts are derived from the previous output via the relation extractor: \(\mu_{\mathcal{F}}(\cdot)\). Here, \(\mathcal{F}\) parameterizes a set of relational functions that includes general object relations like {\small\texttt{distance}} and {\small\texttt{speed}}. Formally, we denote this step as \(\mu_{\mathcal{F}}(\Omega_t) = \Gamma_t = \{g_t^k\}_{k=1}^{d_t}\), where \(d_t\) quantifies the relational concepts.

\textbf{Action Selector: $\Gamma_t \xrightarrow{\rho_{\theta_2}} a_t$}
Finally, the action selector, denoted as $\rho_{\theta_2}$,  determines the action, $a_t$, from the relational concepts.
In contrast to the earlier stages where ICBs provided sufficient interpretability, in this stage the action selector itself must be interpretable to enable overall interpretability (\eg by using decision tree or rule set policies).

\subsection{SPACE}
SPACE~\citep{Lin2020SPACE:} is a model architecture based on \acp{VAE} for unsupervised object-oriented scene representation learning (\cf Figure~\ref{fig:SPACE+MOC}). Its latent space is designed to represent location-related information (\ie the object position) of each object, feature-related information (\ie the object visualization) of each object, and background information separately. SPACE is trained using a standard \ac{VAE} reconstruction loss.

\subsection{MOC}
\begin{figure}[h]
    \centering
    \includegraphics[width=1.0\linewidth]{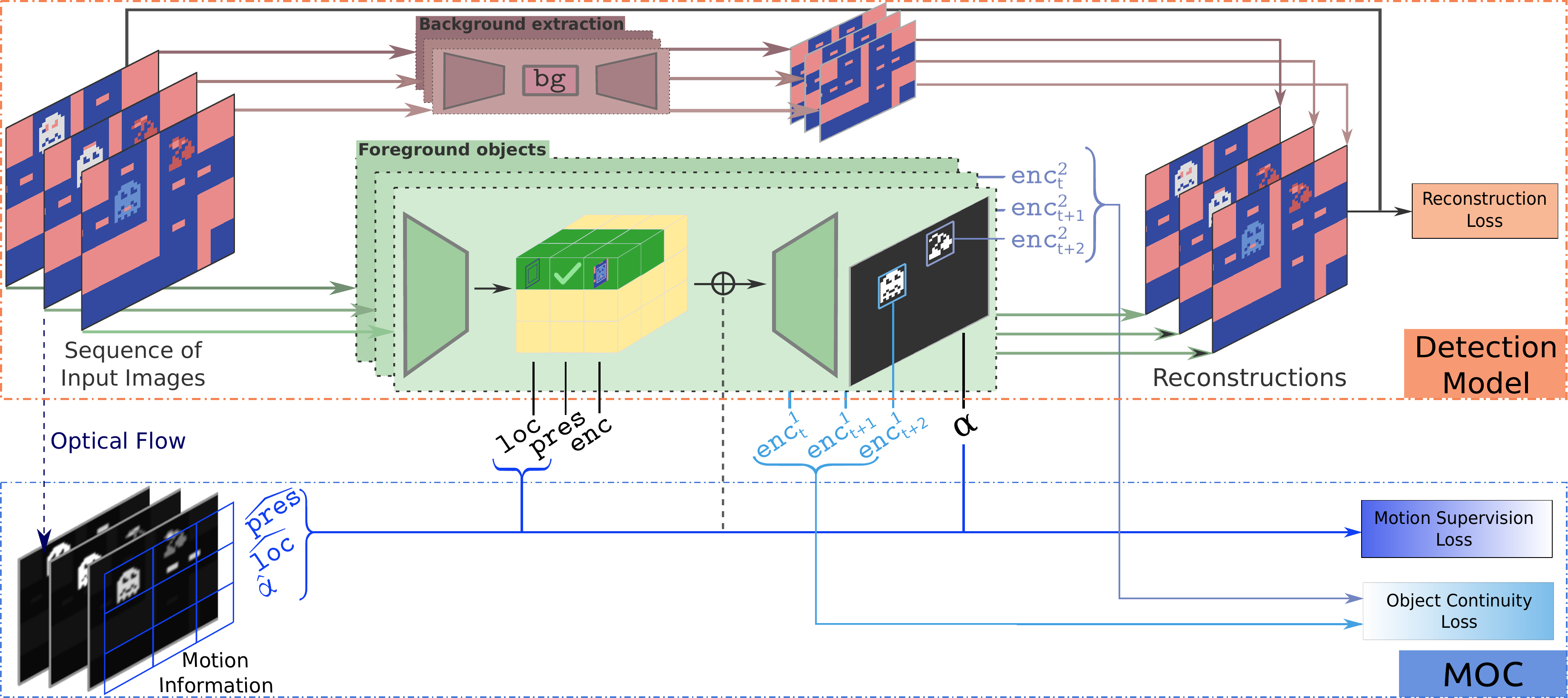}
    \caption{\textbf{Overview of MOC applied to SPACE.} SPACE learns to extract objects using a \ac{VAE} architecture.
    The reconstruction problem is split into a foreground and a background component. The foreground latent space is composed of positional and feature information about the objects in the image. The MOC training scheme adds the motion supervision loss and the object continuity loss to improve a base detection model, SPACE in this case. The motion supervision loss is designed to enhance the localization capabilities by guiding the locations (loc) with motion data of the input image. The object continuity loss is applied to feature encodings (enc) of the objects in consecutive images with the goal of ensuring consistency of the encodings representing the same entity.}
    \label{fig:SPACE+MOC}
\end{figure}

To address the insufficient performance of both object localization and representation learning of SPACE, a follow-up work has added two loss terms within the \acf{MOC} training scheme~\citep{delfosse2023moc}. 
This is an approach that can be applied to any base detection model to improve the object locations (loc) and encodings (enc). The motion supervision loss utilizes additional motion information to improve localization variables (\ie \texttt{loc} and \texttt{pres}). 
The object continuity loss is designed to align object encodings of the same entity across successive frames, and to separate encodings of the different objects. The MOC training scheme is depicted in Figure~\ref{fig:SPACE+MOC}.

\subsection{ECLAIRE}
ECLAIRE~\citep{zarlenga2021efficient} is a rule extraction method for deep neural networks. The input for ECLAIRE is a set of unlabeled training instances \(X = \{x^{(i)} \in \mathbb{R}^m\}_{i=1}^N\) and a pre-trained neural network \(f_{\theta} : \mathbb{R}^m \rightarrow [0, 1]^L\). The function \(f_{\theta}(x)\) outputs a probability distribution over labels in the set \(Y = \{l_1, l_2, \ldots, l_L\}\). 
The output is a set of IF-THEN rules, denoted as \( R_{x \rightarrow \hat{y}} \). These rules are designed to collectively predict the outcome that corresponds to the maximum value in the output vector of the network \( f_\theta \) when subjected to a majority vote given the input \( x \).  A single rule is formalized as follows:

\[
\text{IF} \left( (x_i > v_i) \land (x_j \leq v_j) \land \ldots \land (x_n > v_n) \right) \text{THEN }l_k
\]

In this structure, \( x_i \) represents the \( i \)-th feature of an input instance \( x \), while \( v_i \) is a threshold value determined through the learning process. These rules are composed of premises that are conjunctions of conditions like \( (x_i > v_i) \) or \( (x_i \leq v_i) \).

\section{Implementation}

\begin{figure}[t]
    \centering
    \includegraphics[width=1.\linewidth]{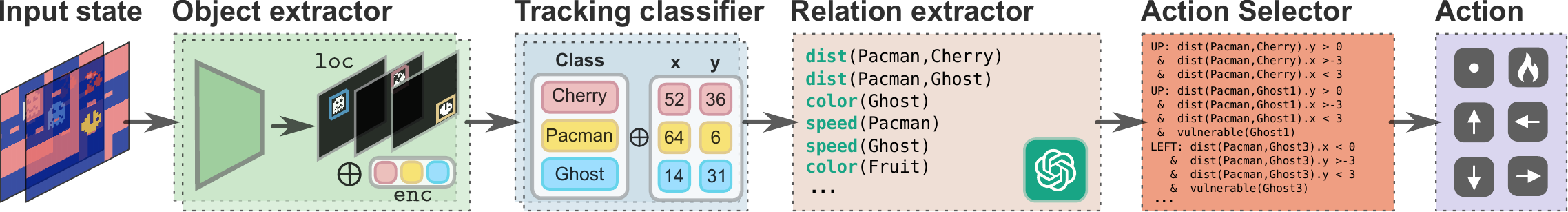}
    \caption{\textbf{Overview of the implementation.} The input images are fed consecutively into the object extractor. The resulting objects represented by location and encoding are classified into a specific object class of the respective game. Also, object identity between consecutive frames is inferred via object tracking. This results in a set of objects with properties, including time-related ones. The relation extractor computes  inter- or intra-object relations. Finally, the action selector decides, which action to take using an interpretable algorithm.}
    \label{fig:DINSA}
\end{figure}

Let us introduce our implementation (\cf Figure~\ref{fig:DINSA}). It uses the SPACE~\cite{Lin2020SPACE:} architecture, trained with the motion and object consistency (MOC) loss of~\cite{delfosse2023moc}. It then uses a k-means classifier to classify each object and a simple object tracking algorithm to infer object identity over consecutive frames. The object-centric space is then augmented with relations and provided to an action selector that uses a set of rules distilled from a neural policy. Let us now detail each component.

\subsection{Object Extractor}
The object extractor receives the last $n$ frames of the game as input. 
It returns the objects of the current frame together with their properties. 
These properties can be time-related, which is why a sequence of images is given as input.

\subsubsection{SPACE+MOC for Object Representation Learning}
This component receives an image as input and produces two key outputs: a bounding box for each detected object and an encoding for each object.
We use the SPACE architecture and enhance it using the MOC training scheme. The desired outputs are obtained from the latent space of the \ac{VAE} architecture of SPACE.

During inference, a set of objects with bounding box and encoding information must be obtained. To this end, only the encoder parts of the foreground module of the SPACE model are required. The other components of SPACE and the MOC training scheme are only needed for training.
The variables  \( z^{pres} \) , \( z^{where} \) and \( z^{what} \) are extracted from the SPACE architecture. The value of \( z^{pres} \) is thresholded, resulting in a binary variable indicating the presence of an object. 
For the cells in which an object is present, the \( z^{where} \) information is transformed into a bounding box and the \( z^{what} \) encoding is saved. 
For implementation details and hyperparameter values, see App.~\ref{app:space+moc-details}.

\subsubsection{Object Classification}
This component is designed to classify objects based on their feature encodings as inputs. In the context of the SPACE+MOC model, these encodings are provided by the $z^{what}$ latent variable. The output of the classification component is a label assigned to each object.

The classifier should be unsupervised in order to not break the overall unsupervised setting.
Similar to~\citep{simeoni2021LOST} and~\citep{Kara2022ImageSU}, we classify the representation of an object based on its distance to the centroids that result from applying k-means clustering to a training set of encodings. A full description of the algorithm can be found in the Appendix at~\ref{app:classifier-details}.

\subsubsection{Object Tracking}
The property {\small\texttt{position history}} is usually required as part of the object representations, as it is essential for computing time-related relational concepts such as {\small\texttt{speed}} in the downstream relation extractor. In order to include this property, the number of image frames $n$ to be contained in the state \(s_t = \{x_i\}_{i=t-(n-1)}^t\) must be set to (at least) two. Moreover, it is necessary to determine which of the localized objects represent the same entity across the sequence of frames. In~\citep{Delfosse2024InterpretableCB}, this information was provided implicitly as part of the ground truth detection via OCAtari. Our solution approach is to use a simple tracking algorithm on top of the single frame localizations by SPACE+MOC. More details are provided in the Appendix at~\ref{app:object-tracking-details}. 

\subsection{Relation Extractor}
The relation extractor uses the extracted objects' properties from the current frame as input, and outputs a vector containing the values of the relational concepts for the detected objects.

The SCoBots framework includes a complete implementation of the relation extractor. In this implementation, 
the number of detectable objects per class are specified in advance, resulting in a set of unique identifiers for potentially detected objects. The relational concepts are then defined relative to these identifiers using straightforward functions, such as Euclidean distance, to generate scalar values for each concept. During inference, detected objects are mapped to an identifier by sorting them based on their proximity to a key object (\eg the player), with excess objects discarded and missing ones assigned zero values. For a more detailed description, see App.~\ref{app:relation-extractor details}.
\subsection{Action Selector}
The action selector component receives the feature vector from the relation extractor and returns an action. The learning phase for our implementation of this component is divided into two steps. First, a neural policy is learned using standard deep \ac{RL} techniques. Then, this policy is transformed into an interpretable representation that uses a set of rules. This transformation is a trade-off between maintaining the policy's similarity and finding a small, easily interpretable set of rules.

\subsubsection{Deep Reinforcement Learning}
This component learns a neural policy based on the relational concepts. This neural policy determines the actions or decisions made in a given context based on the input information. We utilize \ac{PPO}~\citep{Schulman2017ProximalPO} to learn the neural policy. Other RL algorithms that can handle continuous state spaces and discrete action spaces could have also been used. Details on the choice of the hyperparameters are provided in App.~\ref{app:PPO}.

\subsubsection{Policy Distillation via ECLAIRE}
In this step, we transform the neural policy into a rule set policy using ECLAIRE~\citep{zarlenga2021efficient}. Applied to our case, the training instances $X$ in ECLAIRE are the one-dimensional vectors, which the relation extractor provides. The neural network \(f_{\theta}\) is the policy network learnt using the PPO algorithm. $Y$ is the discrete action space of the respective game.

\begin{wrapfigure}[15]{r}{0.5\linewidth}
\vspace{-0.45cm}
\begin{minipage}{0.166\textwidth}
    \centering
    \includegraphics[width=0.98\linewidth]{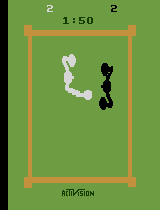}
\end{minipage}\hfill
\begin{minipage}{0.166\textwidth}
    \centering
    \includegraphics[width=0.98\linewidth]{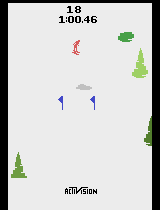}
\end{minipage}\hfill
\begin{minipage}{0.166\textwidth}
    \centering
    \includegraphics[width=0.98\linewidth]{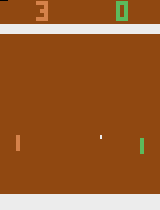}
\end{minipage}
\caption{\textbf{Visualization of the games used in our experimental evaluation.} The Boxing, Skiing, and Pong Atari environments are shown from left to right. While Boxing and Pong were also used for RL evaluation, the difficult credit assignment game Skiing is only used for object detection.}
\label{fig:visualization_of_games}
\end{wrapfigure}

\section{Experimental Evaluation}

In our experiments, we successively evaluate the different components of our \ac{SCoBots}~\citep{Delfosse2024InterpretableCB} instantiation.
First, the object extraction is evaluated. As the relation extractor only applies deterministic functions to object properties, it is not investigated separately. 
Second, the action selector is analyzed including both the preliminary neural policy and the final rule set policy.

The Pong, Boxing and Skiing environments (depicted in Figure~\ref{fig:visualization_of_games}) were used in the experiments for the object extractor. Only Pong and Boxing remained for the action selector experiments, as Skiing is a difficult credit assignment problem, that requires additional techniques to be solved (\cf App.~\ref{app:games}).
The object extractor focuses on the moving objects of the games, only considering the relevant objects for playing the games (\eg not considering scores or the clock). This was realized by applying a filter based on potential detection areas of the moving objects (\cf App.~\ref{app:filtering-objects}).

\subsection{Object Extractor}

In this subsection, we present the evaluation of our object extractor. However, we only included the localization and encoding component plus the classifier, but did not include the object tracker. The scores were calculated relative to \textit{all} ground truth objects and not only relative to the \textit{localized} objects. This allowed us to obtain a better understanding of how the object extractor would behave for the downstream task.

\begin{wrapfigure}[16]{r}{0.5\linewidth}
    \centering
    \vspace{-2cm}
    \includegraphics[width=1.\linewidth]{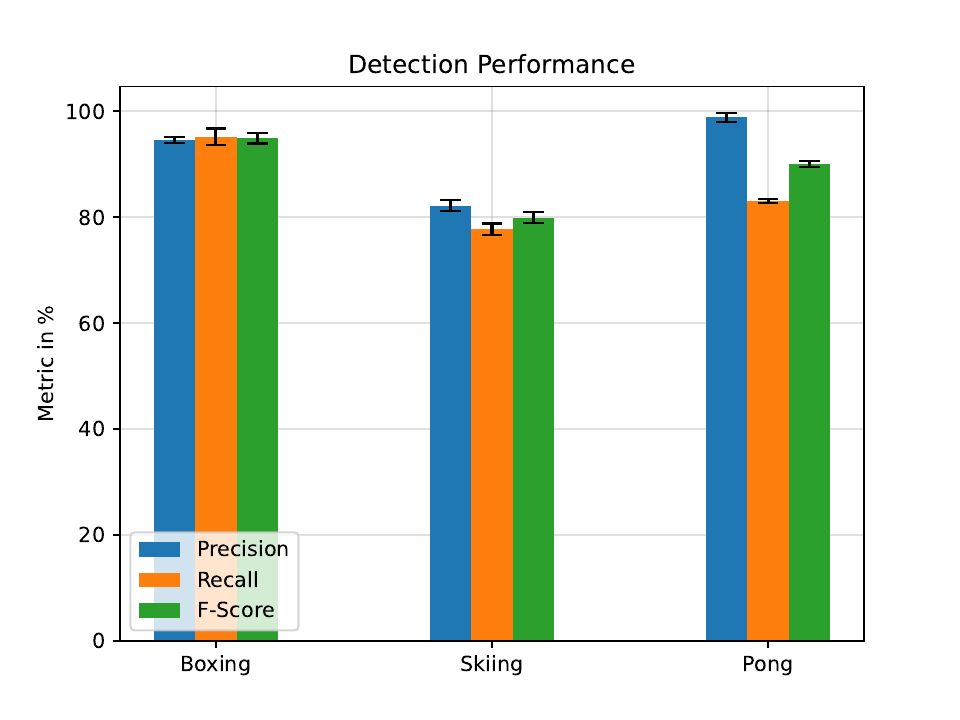}
    \vspace{-0.8cm}
    \caption{\textbf{Object extractors for Boxing and Pong demonstrate high performance, while Skiing shows lower performance.} Precision, recall, and F-score for the combined object extractors of SPACE+MOC and classifier are shown. Results are averages with standard deviations across five SPACE+MOC models trained with different seeds and their corresponding classifiers.}
    \label{fig:f-scores-detection}
\end{wrapfigure}

\begin{figure}[t]
\begin{minipage}{0.333\textwidth}
    \centering
    \includegraphics[width=0.9\linewidth]{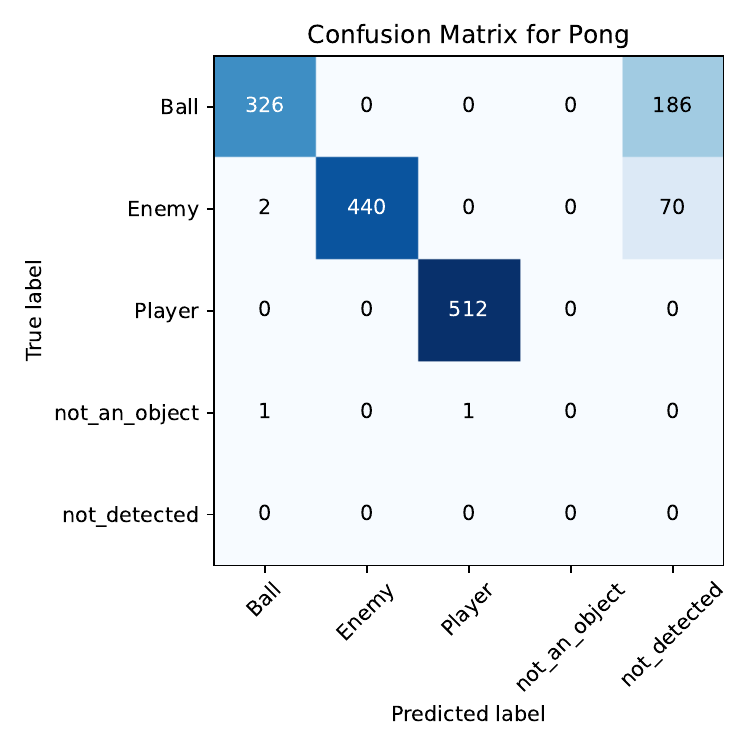}
\end{minipage}\hfill
\begin{minipage}{0.333\textwidth}
    \centering
    \includegraphics[width=0.9\linewidth]{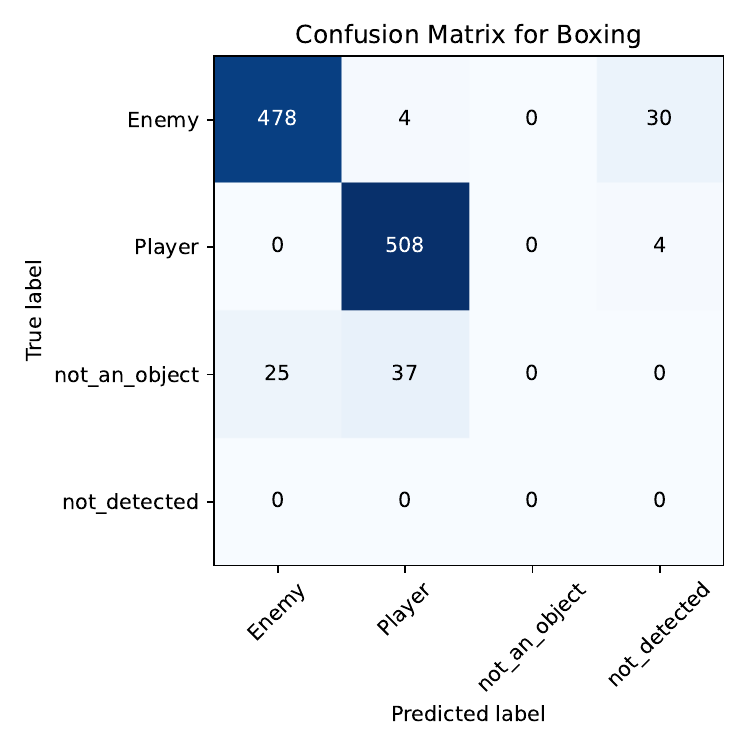}
\end{minipage}\hfill
\begin{minipage}{0.333\textwidth}
    \centering
    \includegraphics[width=0.9\linewidth]{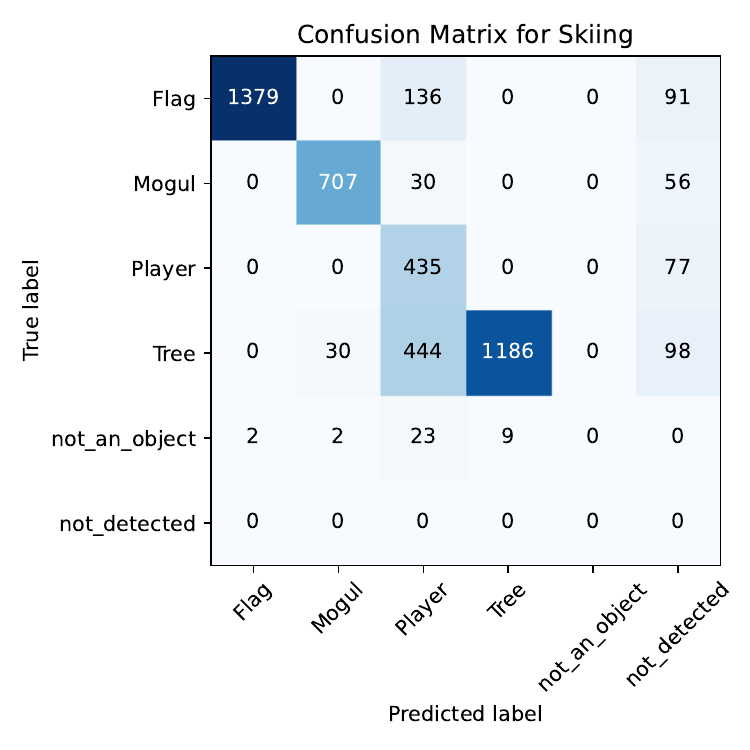}
\end{minipage}
\caption{\textbf{Confusion matrices for each object type per game.} \textit{not\_an\_object} refers to bounding boxes returned by SPACE+MOC that do not correspond to a ground truth object. \textit{not\_detected} refers to objects provided by OCAtari that were not localized by SPACE+MOC.}
\label{fig:conf-matrices-detection}
\end{figure}

Overall, the F-score was the best for Boxing (\cf Figure~\ref{fig:f-scores-detection}). Pong had a slightly lower F-score due to the poor recall for the localization of the ball object. Both are likely to be suitable for the downstream task.
Skiing performed the worst, which can be explained by the poor classifier performance. The average correct detection of only four out of five ground truth objects is unlikely to be sufficient for the downstream task. In particular, when we examined the confusion matrix (\cf Figure~\ref{fig:conf-matrices-detection}), we observed that the detection of the player object was problematic, arguably the most important object. Many trees were classified as the player. This behavior would confuse the downstream RL algorithm.


\newpage
\subsection{Action Selector}
Let us here evaluate the action selection process.

\subsubsection{Deep Reinforcement Learning}
This experiment assesses the performance of \ac{RL} agents equipped with neural policies, with object-centric state input that are provided via the components of the SPACE+MOC object extractor and relation extractor.

The results (\cf Table~\ref{tab:eclaire-experiments}) for Pong reveal that an agent employing the SPACE+MOC object extractor can achieve comparable performance to an agent utilising a ground truth object extractor, provided that the two hidden layer configuration is used . The outcomes of the Boxing experiment indicate that the SPACE+MOC object extractor is too inaccurate for use in competitive object-centric agents. The agents using the SPACE+MOC data achieve poor performance compared to the agents using ground data. The latter agents perform well in particular for both of the unpruned configurations. To summarize, the experiment indicates that the \ac{SCoBots} framework's modular design which enhances interpretability and allows for incremental component upgrades, can come at the cost of error accumulation.

\subsubsection{Policy Distillation}

The goal of this experiment was to ascertain the final performance score for our \ac{SCoBots} implementation. Additionally, we sought to determine the extent to which performance is diminished by extracting the rule set policy from the neural policy. We also aimed to understand how the size of the neural network and the set of relational concepts in the feature vector affect the performance. The results are shown in Table~\ref{tab:eclaire-experiments}.

The findings suggest the benefits of using the pruned and two-layer configuration for distillation via ECLAIRE, although the trends are not entirely clear. Notably, a configuration for a SCoBot using SPACE+MOC input and a rule set policy was identified that achieved a respectable average reward of 14.4 in the game Pong. The best configuration for Boxing using SPACE+MOC input and a rule set policy achieved an average reward of 51.8, although using the unpruned configuration. Overall, the action selector generated interpretable rule set policies with performance nearly matching neural policies under certain conditions.

\begin{table}[t]
    \centering
    
    \setlength{\tabcolsep}{8.5pt} 
    \begin{tabular}{l cccc} 
    \toprule 
    \textbf{Pong} & \multicolumn{2}{c}{\textbf{unpruned}} & \multicolumn{2}{c}{\textbf{pruned}} \\
    \cmidrule(r){2-3} \cmidrule(l){4-5} 
     & \textbf{2 layer} & \textbf{1 layer} & \textbf{2 layer} & \textbf{1 layer} \\ 
    \midrule 
    \textbf{Neural | Ground Truth Objects} & $17.4\pm\scriptsize{1.6}$ & $17.0\pm\scriptsize{2.3}$ & $19.0\pm\scriptsize{1.9}$ & $14.6\pm\scriptsize{1.0}$\\
    \textbf{Rule Set | Ground Truth Objects} & $4.8\pm\scriptsize{5.7}$ & $-15.8\pm\scriptsize{3.1}$ & $15.0\pm\scriptsize{3.0}$ & $10.6\pm\scriptsize{2.4}$\\
    \textbf{Rule Set | SPACE+MOC Objects} & $-7.6\pm\scriptsize{9.2}$ & $-3.2\pm\scriptsize{6.3}$ & $14.4\pm\scriptsize{2.6}$ & $-3.8\pm\scriptsize{3.4}$\\
    \textbf{Neural | SPACE+MOC Objects} & $16.8\pm\scriptsize{1.5}$ & $5.4\pm\scriptsize{3.1}$ & $16.8\pm\scriptsize{2.3}$ & $-1.0\pm\scriptsize{6.6}$\\
    \midrule 
    \textbf{Image Data} & \multicolumn{4}{c}{16.4} \\
    \textbf{Random} & \multicolumn{4}{c}{-20.7} \\
    \textbf{Human} & \multicolumn{4}{c}{9.3} \\ 
    \bottomrule 
    \end{tabular}
    
    \vskip\baselineskip 
    
    \setlength{\tabcolsep}{8pt} 
    \begin{tabular}{l cccc} 
    \toprule 
    \textbf{Boxing} & \multicolumn{2}{c}{\textbf{unpruned}} & \multicolumn{2}{c}{\textbf{pruned}} \\ 
    \cmidrule(r){2-3} \cmidrule(l){4-5} 
     & \textbf{2 layer} & \textbf{1 layer} & \textbf{2 layer} & \textbf{1 layer} \\ 
    \midrule 
    \textbf{Neural | Ground Truth Objects} & $93.0\pm\scriptsize{3.9}$ & $97.0\pm\scriptsize{2.5}$ & $78.8\pm\scriptsize{4.7}$ & $47.4\pm\scriptsize{10.1}$\\
    \textbf{Rule Set | Ground Truth Objects} & $91.2\pm\scriptsize{7.7}$ & $77.4\pm\scriptsize{14.3}$ & $67.2\pm\scriptsize{6.7}$ & $46.0\pm\scriptsize{5.7}$\\
    \textbf{Rule Set | SPACE+MOC Objects} & $51.8\pm\scriptsize{8.2}$ & $9.2\pm\scriptsize{11.5}$ & $37.8\pm\scriptsize{16.8}$ & $38.2\pm\scriptsize{6.0}$\\
    \textbf{Neural | SPACE+MOC Objects} & $65.0\pm\scriptsize{11.3}$ & $56.8\pm\scriptsize{4.4}$ & $39.8\pm\scriptsize{12.9}$ & $38.4\pm\scriptsize{9.3}$\\
    \midrule 
    \textbf{Image Data} & \multicolumn{4}{c}{90.3} \\
    \textbf{Random} & \multicolumn{4}{c}{0.1} \\
    \textbf{Human} & \multicolumn{4}{c}{4.3} \\ 
    \bottomrule 
    \end{tabular}
    
    \caption{\textbf{Overview of PPO and rule extraction experiments results.} Results using PPO with object-centric input from either the SPACE+MOC object extractor or a ground truth object extractor based on OCAtari and results using rule set policies with input from either the SPACE+MOC object extractor or a ground truth object extractor based on OCAtari. Average rewards with standard deviations across five differently seeded evaluation episodes. PPO agents using image input~\citep{Delfosse2024InterpretableCB}, random agents, and human scores~\citep{Mnih2015HumanlevelCT} are provided for comparison. Pong and Boxing have a maximum achievable reward of 21 and 100 respectively.}
    \label{tab:eclaire-experiments} 
\end{table}

\section{Limitations}

The current approach relies on strong assumptions about the training environment, such as the availability of training images showing all object variations and motion data from optical flow estimation. The first aspect can be problematic in environments where objects appear only after certain thresholds, requiring pre-trained agents to collect data. Furthermore, it is uncertain whether valuable motion data can be obtained in more complex scenarios than Atari games.

Currently, the extracted properties only concern the location and the class of the objects. More advanced properties such as the {\small\texttt{orientation}} of an object are currently not extracted, even though they can be highly relevant in some Atari games (\eg in Skiing).

The rule set representation of the policy lacks interpretability due to a large number of generated rules, complex premises with many terms and the fact that the premises of multiple rules with conflicting outputs can be satisfied for the same input data point.


\section{Future Work}

Replacing the implementation of the object extractor with unified object detection and tracking methods (\eg YOLO~\citep{Redmon2015YouOL,Wang2024YOLOv9LW}) could be promising, although this would lead to a limited set of properties. Keeping the multi-step implementation of the object extractor, alternative models to SPACE+MOC could be investigated such as SlotAttention~\citep{Locatello2020ObjectCentricLW} or CutLER~\citep{Wang2023CutAL}.
In addition, leveraging the sequential nature of the images, beyond the MOC framework, could improve the robustness and reliability of the object extractor (\eg by incorporating a Kalman filter in the object tracking step). Another avenue for further investigation is enhancing the object extractor's capacity to identify additional properties.


Further investigation into the action selector's interpretability, including tuning ECLAIRE's hyperparameters for simpler rule sets, analyzing actions from a human perspective, and exploring alternative policy distillation methods, could be promising.

Expanding the \ac{SCoBots} framework to a wider variety of games and three-dimensional environments could provide valuable insights, particularly by testing the effectiveness of optical flow estimation for motion supervision, and is essential for advancing toward real-world applications.

\section{Related Work}
\subsection{Explainable and Interpretable Reinforcement Learning}
Explainable Reinforcement Learning (XRL) is a prominent area within the field of \ac{XAI}, focusing on giving human insights into the decision-making processes of AI agents. Key publications such as~\citep{Milani2023ExplainableRL, Dazeley2021ExplainableRL, Krajna2022ExplainabilityIR,Vouros2022ExplainableDR, Qing2022ASO} have extensively reviewed the field of XRL. These works propose frameworks for categorization, highlight complexities, and emphasize ongoing issues that require resolution.

\ac{SCoBots} can be categorized as an \textit{intrinsic} approach as it directly allows humans to grasp how the model reaches its predictions without requiring any additional computation as \textit{post-hoc} approaches would~\citep{Krajna2022ExplainabilityIR}. However, they are only truly an intrinsic approach if an appropriate action selector is chosen, as in our case a rule set policy.
The explanations provided by a SCoBot are usually \textit{local} as they are focused on a single input instance, in contrast to global explanations, which would enable understanding the overall input-output behavior~\citep{Krajna2022ExplainabilityIR}. Again, the choice of the action selector is the main determining factor. Our instantiation via a rule set policy leads to local explanations.
The \ac{SCoBots} framework falls under the \textit{feature importance method} category as defined by~\citep{Milani2023ExplainableRL}, emphasizing the identification of crucial input features that influence decisions.
\ac{SCoBots} provide \textit{zero-order explanations} according to the framework by~\citep{Dazeley2021ExplainableRL}, focusing on the agent's immediate response to inputs. 
The \ac{SCoBots} approach aligns with \textit{model explaining} as outlined in~\citep{Qing2022ASO}, with the focus on elucidating the model's rationale.
According to the categorization by~\citep{10.1007/s10994-024-06543-w}, \ac{SCoBots} learn \textit{Symbolic Representations} and use \textit{Object-Recognition}.
That categorization focuses solely on intrinsic approaches, for which the authors reserve the term \textit{interpretable}.
With regard to \textit{Interpretable Decision-Making}, the \ac{SCoBots} framework itself does not fall into a specific category. However, our use of policy distillation via rule extraction classifies as an \textit{Indirect Approach} in the subcategory \textit{Decision Trees and Variants}.

\subsubsection{Policy Distillation}
Policy distillation~\citep{Rusu2015PolicyD, czarnecki2019distilling}, a specialized form of knowledge distillation~\citep{Hinton2015DistillingTK}, involves training a highly performing teacher model and subsequently distilling this knowledge into a simpler student model. By selecting an appropriate student model architecture with sufficient constraints on complexity, an interpretable yet still performing model can be derived. Policy distillation is particularly advantageous when the desired final model architecture is challenging to create independently due to less performing optimization algorithms.

One notable implementation of this concept is VIPER~\citep{Bastani2018VerifiableRL}, which utilizes imitation learning to extract decision tree policies from a neural policy and Q-function. MoËT~\citep{Vasic2019MoTIA} extends this approach by incorporating a mixture of expert trees. MAVIPER~\citep{Milani2022MAVIPERLD} adapts VIPER to a multi-agent setting. Furthermore, approaches from the field of rule extraction can be used for policy distillation. These approaches transform the teacher model into a set of explicit IF-THEN rules. ECLAIRE~\citep{zarlenga2021efficient} exemplifies this approach and is used in our instantiation of the \ac{SCoBots} framework. 
NUDGE~\citep{delfosse2024interpretable} is an approach from the domain of Neural Logic Reinforcement Learning~\citep{Jiang2019NeuralLR}.
This approach uses a neural policy to guide the search for a promising rule set. Furthermore, it can leverage the Q-function, if available, to initialize the critic in its actor-critic RL algorithm.
EXPIL \cite{sha2024expil} integrates predicate discovery to reduce the reliance on experts, and BlendRL~\citep{shindo2024blendrl} uses a mixture of deep and logic policies to overcome the potential lack of available concepts.
Another approach is INTERPRETER~\citep{kohler2024interpretable}. This method creates interpretable programs by using the concept of oblique decision trees.

\subsection{Object Representation Learning}
If the properties obtained by the object extractor only include the location and object class, as in our experiments, an object detection system suffices. In this case, models such as CutLER~\citep{Wang2023CutAL} and LOST~\citep{simeoni2021LOST} could be considered. These have demonstrated strong performance by leveraging features obtained through self-supervised learning combined with vision transformers~\citep{Caron2021EmergingPI}.
However, ultimately the goal is to extract more properties, hence our choice to use an approach from the field of object representation learning.
SPACE~\citep{Lin2020SPACE:} follows a line of work that started with AIR~\citep{NIPS2016_52947e0a}. AIR introduced a sequential attention mechanism that iteratively attends to and infers objects in a scene, using a recurrent neural network to propose object regions and a \ac{VAE} to generate object representations. SPAIR~\citep{Crawford2019SpatiallyIU} modified by introducing spatial invariance, utilizing a grid-based attention mechanism to enhance computational efficiency.
Other approaches that operate on the pixel level, rather than with bounding boxes, include with Tagger~\citep{Greff2016TaggerDU} and NEM~\citep{NIPS2017_d2cd33e9}. Notably, some works have demonstrated the ability to generate representations with dimensions that can be associated with specific features of the objects (\eg color, shape)~\citep{greff2019multi, Burgess2019MONetUS, engelcke2020genesis}. Recently, approaches have emerged that employ the concept of optical flow to benefit from the consecutive nature of the images~\citep{Lao2023DividedAU, kipf2022conditional}, as does \acs{MOC}~\citep{delfosse2023moc}, which is used in our work together with SPACE.
In supervised settings, automatic concept finding to extend the object-centric representations has been developed for explanation~\cite{friedrich2023one, stammer2023learning, stammer2024neural} or automatized with lambda-calculus-based concepts~\cite{wust2024pix2code}.
Interpretable concepts can also be revised by experts in case of misalignment~\cite{steinmann2023learning, Delfosse2024InterpretableCB}. Related to RL, concept revision has also been used in the domain of time series~\cite{kraus2024right}.

\section{Conclusion}
We instantiated the \ac{SCoBots} framework~\citep{Delfosse2024InterpretableCB} and successfully demonstrated its application to Atari games. The \ac{SCoBots} used a trained object detection component instead of the ground truth detection used in previous work. In the process, we explored several critical areas, including object representation learning, which involves simplifying a scene into an object-centric representation, and object-centric RL, which focuses on learning policies based on the representations of objects. Additionally, we covered policy distillation by applying rule extraction, which transforms a neural policy into a more interpretable rule set policy. Through this work, we hope to contribute to the improvement of interpretability in RL. We believe that by improving interpretability, RL agents can be analyzed and designed more successfully, which can facilitate addressing pervasive challenges in the field of RL. We also identified promising future research directions that could further enhance the framework's potential for advancing the field.

\clearpage


\vskip 0.2in
\bibliography{main}
\bibliographystyle{unsrt}

\clearpage

\appendix
\section{Appendix}
This appendix presents supplementary information on our research experiments and results. It includes details on the data set used, model configurations, and evaluation metrics. 

\subsection{Object Extractor}

\subsubsection{Data Set}

The data set was generated using OCAtari~\citep{Delfosse2023OCAtariOA}. For generation, a random agent was used. A total of 2048 training, 128 validation, and 128 test image sequences were collected, each comprising four consecutive frames. The reason for collecting four consecutive frames was that consecutive images are required to calculate the object continuity loss. For each frame, the ground truth object detections were also stored. A deliberate pause of at least 16 steps between sequences was implemented to ensure data diversity. The frames were downscaled to 128x128 pixels for compatibility with the SPACE~\citep{Lin2020SPACE:} model.

\textbf{Games}
\label{app:games}

Pong, Boxing and Skiing were used in the experiments for the object extractor. These games cover different levels of difficulty for the object extractor. Pong and Boxing only contain a very small number of objects in each frame. The shapes of the objects in Boxing are more complex and can vary in shape, depending on whether the boxers are punching or not. Skiing contains more objects and the visual appearance of objects can vary within the same object class. An image of each game is depicted in Figure~\ref{fig:visualization_of_games}.
In the experiments for the action selector, only Pong and Boxing were used. Skiing is too challenging even for end-to-end deep \ac{RL} approaches due to its extremely delayed reward signal.

\textbf{Optical flow}
\label{app:optical-flow}

The optical flow method involved the collection of a mode image based on 100 images. A manual check was conducted to ensure that the background of the respective game was accurately represented. The optical flow and motion data were then calculated using the background subtraction approach outlined in~\citep{delfosse2023moc}.

\subsubsection{SPACE+MOC Details}
\label{app:space+moc-details}
In general, the hyperparameters from~\citep{delfosse2023moc} were reused with a few modifications (\cf Table~\ref{app:spacemocparams}). Overall, the hyperparameters related to SPACE are largely consistent with those presented in the original work~\citep{Lin2020SPACE:}.

\label{app:dynamic-scheduling}
The MOC loss was applied using the dynamic scheduling approach proposed in~\citep{delfosse2023moc}. This approach dynamically balances the motion supervision loss and the object continuity loss depending on the ability to correctly localize the bounding boxes. For more details, please refer to to~\citep{delfosse2023moc}. In contrast to~\citep{delfosse2023moc}, no bootstrapping of the decoder with the help of the motion information was used. The training was repeated using five different seeds, and the resulting data was used to calculate the mean and standard deviation for the metrics.

\begin{table}[ht!]
\centering
\begin{tabular}{|m{6cm}|m{4cm}|}
\hline
\textbf{Parameter} & \textbf{Value} \\
\hline
batch size & 16 \\
\hline
gradient steps & 5000 \\
\hline
\hline
Foreground / Background lr & $3 \cdot 10^{-5} / 10^{-3}$ \\
\hline
$z_{\text{pres}}$ prior probability & $0 \rightarrow 5000 : 0.1 \rightarrow 10^{-10}$ \\
\hline
$z_{\text{scale}}$ prior mean & $0 \rightarrow 5000 : -2 \rightarrow -2.5$ \\
\hline
$z_{\text{scale}}$ prior std & $0.1 \cdot \mathbf{I}$ \\
\hline
$z_{\text{shift}} / z_{\text{what}} / z_{\text{depth}}$ priors & $\mathcal{N} (0, \mathbf{I}) / \mathcal{N} (0, \mathbf{I}) / \mathcal{N} (0, \mathbf{I})$ \\
\hline
$\tau$ (gumbel-softmax-temperature) & 2.5 \\
\hline
Foreground / Background stds & 0.2 / 0.1 \\
\hline
Background Components & 3 \\
\hline
Grid Size & 16 \\
\hline
fixed $\alpha$ \& boundary loss & removed \\
\hline
\hline
Motion Kind & Mode \\
\hline
$\eta$ & 0.5 \\
\hline
$\lambda_{\alpha}/\lambda_{\text{pres}}/\lambda_{\text{where}}$ & 100 / 1000 / 10000 \\
\hline
$\lambda_{\text{guid}}$ & $0 \rightarrow 3000 : 1.0 \rightarrow 0.0$ \\
\hline
$\beta_{\text{mismatch}}, \beta_{\text{underestimation}}$ & 0.1, 1.25 \\
\hline
 ${\beta}_{\text{differ}}$&5\\\hline
 $\lambda_M$/$\lambda_{OC}$&100 / 10\\\hline
\hline
\multicolumn{2}{|c|}{Deviations - Boxing:} \\
\hline
\( z_{\text{scale}} \) prior & -1 \\
\hline
\multicolumn{2}{|c|}{Deviations - Pong:} \\
\hline
\( \beta_{\text{underestimation}} \) & 1.5 \\
\hline
\end{tabular}
\caption{SPACE and MOC shared parameter values for training, SPACE base parameter values, MOC parameter values, Deviations from SPACE+MOC base parameters for Pong and Boxing as in~\citep{delfosse2023moc}}
\label{app:spacemocparams}
\end{table}

\textbf{Filtering SPACE+MOC Localizations}

\label{app:filtering-objects}
The evaluation of the object extractor was focused on the moving objects of the games, only considering the relevant objects for playing the games. Consequently, static objects that are also provided via OCAtari were removed from the ground truth data (\eg the clock object). Furthermore, the predicted objects by the model for localization underwent filtering based on potential detection areas of the moving objects. Practically, this was implemented by discarding localized objects that have bounding box coordinates clearly outside of the areas where moving objects can appear in the game. For the games that were considered in our experiments, this approach is sufficient because non-moving objects can only be found outside of this area. The same approach was also used in~\citep{delfosse2023moc}.
\begin{table}
    \centering
    \begin{tabular}{|c|c|} \hline 
         \textbf{Game}& \textbf{Rule}\\ \hline 
         Boxing& $0.148 < y_{min} \land y_{max} < 0.859$\\ \hline 
         Pong& $0.164 < y_{max} \land 0.031 < y_{min}$\\ \hline 
 Skiing&True\\ \hline
    \end{tabular}
    \caption{Region filters for SPACE+MOC localizations}
    \label{tab:filtering-localizations}
\end{table}
The rule for Pong differs from the rule used in~\citep{delfosse2023moc}.

\subsubsection{Classifier Details}
\label{app:classifier-details}
The creation of the classifier involves multiple steps.
First, k-means clustering is performed to obtain the centroids. For each Atari game, the value for $k$ is given by the number of object classes as specified in OCAtari~\citep{Delfosse2023OCAtariOA}.
Second, descriptive labels are assigned to the obtained centroids, since the k-means clustering only returns enumerated class labels. For this, a k-nearest neighbors classifier is used, where $k$ does not refer to the same number as in k-means. The initialization is based on object encodings extracted by SPACE+MOC from a small image data set. These objects have been assigned a descriptive label based on the object names of the ground truth detections provided by OCAtari. The descriptive labels for the centroids are finally assigned via the k-nearest neighbors classifier. It is important to note that this approach is not entirely in accordance with the unsupervised setting, but the supervised data is only used for assigning names to the classes.
Third, the final classifier is given by a 1-nearest neighbor classifier initialized only with the centroids.

In k-means clustering, the parameter k was set to the number of relevant object types, and only one run was conducted with random initialization of the centroids. For determining descriptive labels for the centroids, the k-nearest neighbors classifier was employed with k set to 24. In all three steps, the Euclidean distance was utilized as the distance metric.

The classifier was trained using latent variables extracted from a SPACE+MOC model, which was applied to the validation set images. The model used was the SPACE+MOC model from the localization experiments with the highest F-score. The training set was not used to avoid data leakage. Only the first of each sequence of consecutive images was used, as the images and extracted latents for images in a sequence would be too similar. For the purposes of testing, the test set was employed, with the remaining aspects left unchanged.

\subsubsection{Tracking Algorithm Details}
\label{app:object-tracking-details}
In the initial pass of the tracking algorithm through a video frame, each detected object is added to a tracking list. Each object is identified by the bounding box that encapsulates it, and at this stage, there are no previous tracks to compare against, so all detections are treated as new objects.
From the second frame onwards, the algorithm calculates the distances between the centroids of the currently tracked objects and the centroids of the new detections.
The matching scores, derived from the distance calculations, are used to determine which new detections correspond to which existing tracks. A detection is assigned to the closest tracked object, thereby ensuring continuity in tracking.
New detections that do not closely match any current track—either because they are too far from existing tracks or are only the second-best match—are initiated as new tracks. This step accounts for new objects entering the scene.
Conversely, objects that have been previously tracked but do not find a match in the new detections are removed from the tracking list. This addresses objects that leave the scene or become occluded.

The algorithm's simplicity can result in failure when objects cross paths or even overlap for multiple consecutive frames. While the former may result in incorrect features for a single frame, the latter can lead to significant issues. The current implementation is based on~\citep{multiobjtracker_amd2018}. 

\subsubsection{Combined Object Extractor Experiments}
The results were generated using the SPACE+MOC models with the highest F-score from five differently seeded runs, in conjunction with a corresponding classifier trained based on the encodings of that model.

The utilized F-score metric is a well-established and widely accepted standard. However, the evaluation of the localizations is heavily reliant on the manner, in which the localized and ground truth bounding boxes are assigned to one another. Therefore, further details on this are provided below.

\textbf{How to match bounding boxes?}

\textbf{Pairwise Matching Scores} -  How to measure the similarity between a predicted and a ground truth bounding box? \label{app:eval-localization-considerations}

The authors of~\citep{delfosse2023moc} argue that the intersection over union (IoU) metric, which is arguably the most common metric in this context, is not ideal for assessing performance for the downstream task of RL for Atari games. This is due to the architecture of SPACE, which tends to return bounding boxes of similar sizes for a game. Consequently, in games such as Pong where the ball and player objects differ in size, at least one of the object classes will have a bounding box that is either too large or too small. This will result in low IoU scores. However, since the objects are (mostly) of constant size in the games that we consider, the downstream RL algorithm can implicitly infer the size. In fact, the object-centric input representation returned by the relation extractor only contains the center coordinates of the objects in our implementation. As an alternative to IoU, the authors of~\citep{delfosse2023moc} introduce the center divergence metric. This metric focuses on the distance of the center coordinates of the bounding boxes. For details, we refer to~\citep{delfosse2023moc}. The center divergence metric was employed in the calculation of precision, recall, and the derived F-score. 

\textbf{Matching} - How to assign predicted bounding boxes to ground truth bounding boxes based on the pairwise matching scores?

The first condition is that the matching score of a predicted bounding box to a ground truth bounding box must exceed a specified threshold. We employed a threshold of 0.5 for the matching score computed via center divergence. In addition to the first condition, further conditions are required because two predicted bounding boxes might have a sufficient matching score with the same ground truth bounding box or a single predicted bounding box might have a sufficient matching score with two ground truth bounding boxes. Due to the relatively simple nature of the images, we applied a straightforward greedy approach. The predicted bounding boxes that met the first condition were sorted by confidence and then were assigned greedily to the ground truth bounding boxes. In more complex scenarios, the use of Hungarian matching is advisable.

\textbf{Joint treatment of Localization and Classification}

In the event that a localized object cannot be matched with a ground truth object, it is labeled as "no object." Conversely, if a ground truth object does not correspond to any localized object, it is designated as "not detected." This classification system allows for a more nuanced understanding of the detections.

In order to evaluate the performance of the object extractor, we calculated the precision, recall, and F-score. It is necessary to be precise with the definition of these metrics. This is due to the two-step approach in which the classifier, in the second step, can receive localized objects that do not have a ground truth object associated with them. At the same time, the classifier does not necessarily receive all of the ground truth objects.
Our calculated precision score addresses the following question: What proportion of all the detected objects corresponds to a ground truth object and is correctly labeled? In other words, when looking at the confusion matrix (\cf Figure~\ref{fig:conf-matrices-detection}), the "not\_detected" column is removed, and then the micro precision is calculated. Analogous to the precision score, our recall score answers the following question: What proportion of all the ground truth objects has been localized and has received the correct label? This is achieved by removing the "not\_an\_object" row from the confusion matrix and then calculating the micro recall.

\subsection{Relation Extractor}
\label{app:relation-extractor details}
\textbf{Algorithmic description}

The relational concepts are specified in advance. To this end, the number of objects of each class to be considered must be specified at first. Objects of the same class are given unique identifiers by enumeration. Then, the relational concepts to be computed based on these objects are defined. A relational concept is characterized by the input objects and the relational function applied to their properties. The relational functions themselves are straightforward (e.g., calculating Euclidean distance). This approach specifies clear computational instructions that yield a single scalar value for each defined relational concept. Each relational concept is assigned a fixed position in the output vector.

During inference, the detected objects must be mapped to the enumerated objects from the predefined computational instructions. For this purpose, the detected objects are sorted in ascending order based on their distance to the player object. The player object is defined as the object controlled by the agent (e.g., the right paddle in Pong). Objects within each class are then enumerated according to this order. If the object extractor detects more objects for a class than specified in advance, those with identifiers exceeding the maximum number are discarded. Conversely, if a predefined object of a class remains unassigned due to fewer detected objects, the relational concepts dependent on this object are assigned a scalar value of zero.

\textbf{Selection of relational concepts}
\begin{itemize}
    \item  Full set: All relational concepts as detailed in Table 3 of~\citep{Delfosse2024InterpretableCB} were included. This set considers all possible combinations of objects for n-ary relations, including symmetric ones like distance, where both directions (\eg d(Ball1, Player1) and d(Player1, Ball1)) are accounted for, resulting in six combinations. For the {\small\texttt{linear trajectory}} relation, combinations with the object itself are included, \eg leading to nine combinations for three objects. 
    \item Pruned set: The pruned set of relational concepts only includes the subset of relational concepts that are presumed to be relevant for the game as listed in Table 4 of~\citep{Delfosse2024InterpretableCB}. The selection of relational concepts is based on human understanding of the games.
\end{itemize}
Both the full set and the pruned set stem from~\citep{Delfosse2024InterpretableCB}.

\subsection{Action Selector}
\label{app:properties-and-relations}

\subsubsection{Experiment Design}
We designed the experiments to encompass different levels of complexity of the model for the neural policy. This was done with the rationale that the rule set extraction would probably work more reliably the simpler the \ac{NN} structure and the policy is. Therefore, we varied the depth of \ac{NN} architecture in the \ac{PPO}~\citep{Schulman2017ProximalPO} algorithm. The \acp{NN} were instantiated with either 1 or 2 hidden layers of 64 neurons. Additionally, the set of relational concepts was varied. Either the full set of relations or a pruned version was used (\cf App.~\ref{app:relation-extractor details}). 


Different versions for the Atari games are available in OCAtari via the underlying Gymnasium library~\citep{towers_gymnasium_2023}. Following the recommendations of~\citep{Machado2018RevisitingTA} and for better comparability with~\citep{Delfosse2024InterpretableCB}, v5 of the games was used. 

\subsubsection{Deep Reinforcement Learning Experiments}
\label{app:PPO}
The hyperparameters for PPO~\citep{Schulman2017ProximalPO} mostly stem from~\citep{Delfosse2024InterpretableCB}, with a few minor exceptions. Specifically, 10M frames were used instead of 20M, and only a single training seed was used. In contrast to~\citep{Delfosse2024InterpretableCB}, the policy and value networks were initialized with either 1 or 2 hidden layers of size 64. One important aspect to point out is that the advantages in the experiments were normalized as in~\citep{Delfosse2024InterpretableCB}.
The existing implementation of PPO in the package stable-baselines3~\citep{stable-baselines3} was used.

\subsubsection{Policy Distillation Experiments}
For ECLAIRE~\citep{zarlenga2021efficient}, we retained the default hyperparameter settings. The main hyperparameter that could be tuned for better performance is $\mu$. This parameter specifies the minimum number of samples that are required for C5.0 to have before splitting a node. 
As mentioned in the Method chapter, the training instances must be diverse enough. For this purpose, we used the PPO policy, with a random action taken in 25\% of cases. A total of 50,000 training instances were collected using this strategy, from which the rules were extracted.
For inference, the conclusion of the rule with the highest confidence was chosen among all the rules whose premises are satisfied by the input data point.

\end{document}